\documentclass[11pt,a4paper]{article}
\usepackage[hyperref]{eacl2021}
\usepackage{times}
\usepackage{arabtex}
\usepackage{latexsym}

\usepackage{float}
\restylefloat{table}
\usepackage{microtype}
\aclfinalcopy 
\usepackage{enumitem}

\title{Automatic Difficulty Classification of Arabic Sentences}

\author{Nouran Khallaf, Serge Sharoff \\
  School of Languages, University of Leeds\\
  Leeds, LS2 9JT, United Kingdom
  \texttt{mlnak,s.sharoff@leeds.ac.uk} 
}
\date{}

\begin{document}
\maketitle
\begin{abstract}
In this paper, we present a Modern Standard Arabic (MSA) Sentence difficulty classifier, which predicts the difficulty of sentences for language learners using either the CEFR proficiency levels or the binary classification as simple or complex. We compare the use of sentence embeddings of different kinds (fastText, mBERT , XLM-R and Arabic-BERT), as well as traditional language features such as POS tags, dependency trees, readability scores and frequency lists for language learners.
Our best results have been achieved using fined-tuned Arabic-BERT. The accuracy of our 3-way CEFR classification is F-1 of 0.80  and 0.75 for Arabic-Bert and XLM-R classification respectively and 0.71 Spearman correlation for regression.
Our binary difficulty classifier reaches F-1 0.94 and F-1 0.98 for sentence-pair semantic similarity classifier.

\end{abstract}

\section{Introduction}

In the last century, measuring \emph{text readability (TR)} has been undertaken in education, psychology, and linguistics. There appears to be some agreement that TR is the quality of a given text to be easy to comprehend by its readers in adequate time with reasonable effort \citep{cavalli2018arabic}. Research to date has tended to focus on assigning readability levels to whole text rather than to individual sentences, despite the fact that any text is composed of a number of sentences, which vary in their difficulty \citep{schumacher2016predicting}. Assigning readability levels for a text is a challenging task and it is even more challenging on the sentence level as much less information is available. Also, the sentence difficulty is influenced by many parameters, such as, genre or topics, as well grammatical structures, which need to be combined in a single classifier. Difficulty assessment at the sentence level is a more challenging task in comparison to the better researched text level task, but the availability of a readability sentence classifier for Arabic is vital, since this is a prerequisite for research on \emph{automatic text simplification} (ATS), i.e. the process of reducing text-linguistic complexity, while maintaining its meaning \citep{saggion2017automatic}.

We focus here on experiments aimed at measuring to what extent a sentence is understandable by a reader, such as a learner of Arabic as a foreign language, and at exploring different methods for readability assessment.  The main aim of this paper lies in developing and testing different sentence representation methodologies, which range from using linguistic knowledge via feature-based machine learning to modern neural methods.

In summary, the contributions of this paper are:
 \begin{enumerate}[noitemsep]
  \item We compiled a novel dataset for training on the sentence level; 
  \item We developed a range of linguistic features, including POS, syntax and frequency information;
  \item We evaluated a range of different sentence embedding approaches, such as fastText, BERT and XLM-R, and compared them to the linguistic features;
  \item We cast the readability assessment as a regression problem as well as a classification problem;
  \item Our model is the first sentence difficulty system available for Arabic.
\end{enumerate}

\section{Corpora and Tools}
\subsection{Dataset One: Sentence-level annotation}
This dataset was used for Arabic sentence difficulty classification. We started building our own dataset by compiling a corpus from three available source classified for readability on the document level along with a large Arabic corpus obtained by Web crawling.

The first corpus source is the reading section of the \textbf{Gloss}\footnote{https://gloss.dliflc.edu/} Corpus developed by the Defense Language Institute (DLI). It has been treated as a gold standard and used in the most recent studies on document level predictions \citep{forsyth2014automatic,saddiki2015text,nassiri2018arabic,nassiri2018modern}. Texts in Gloss have been annotated on a six level scale of the Inter-Agency Language Roundtable (IL ), which has been matched to the CEFR levels according to the schema introduced by \citep{tschirner2015assessing}. Gloss is divided according to the four competence areas (lexical, structural, socio-cultural and discursive) and ten different genres (culture, economy, politics, environment, geography, military, politics, science, security, society, and technology). 

The second corpus source is the \textbf{ALC} , which consists of Arabic written text produced by learners of Arabic in Saudi Arabia collected by \cite{alfaifi2013arabic}. Each text file is annotated with a proficiency level of the student. We mapped these student proficiency levels to CEFR levels.

Our third corpus source comes from textbook \textbf{"Al-Kitaab fii TaAallum al-Arabiyya"} \citep{brustad2015kitaab} which was compiled from texts and sentences from parts one and two of the third edition but only texts from part three third edition.  This book is widely used to teaching Arabic as a second language. These texts were originally classified according to the American Council on the Teaching of Foreign Languages (ACTFL) guidelines which we mapped to CEFR levels.  

As these corpora have been annotated on the document level and not on the sentence level, we assigned each sentence to the document level in which it appears, by using several filtering heuristics, such as sentence length and containment, as well as via re-annotation through machine learning, see the dataset cleaning procedure below.

A counterpart corpus of texts not produced for language learners in mind is provided by I-AR, 75,630 Arabic web pages collected by wide crawling \citep{sharoff06ijcl}.  A random snapshot of 8627 sentences longer than 15 words was used to extend the limitations of C-level sentences coming from corpora for language learners.

Table ~\ref{Data-Set1} shows distribution of the number of used sentences and tokens per each Common European Framework of language proficiency Reference [CEFR] Level. In principle we have data for 5-way (A1, A2, B1, etc), 3-way (A, B or C) and binary (A+B vs C) classification tasks, but here in this presentation, we focus on the 3-way and binary (simple vs complex) classification tasks.

\begin{table}[t]
\centering
\begin{tabular}{lcccc}
{CEFR} & \multicolumn{2}{c}{\textbf{Old}}  &\multicolumn{2}{c}{\textbf{New}} \\
\hline
 & S & T & S & T \\
 \hline
A & 8661 & 187225 & 9030 & 195343 \\
B & 5532 & 126805 & 5083 & 117825 \\
C & 8627 & 287275 & 8627 & 287275 \\
\hline
Total & 22820 & 601305 & 22740 & 600443\\
\hline
\end{tabular}
\caption{\label{Data-Set1}(S)sentences and (T)tokens available per each CEFR Level in the two versions of the corpus}
\end{table}

\paragraph{\emph{Dataset cleaning:}}
In our initial experiments we noticed unreliable sentence-level assignments in the training corpus. Therefore, we decided to improve the quality of the training corpus by an error analysis strategy introduced by \citet{di2014multiple}, which is based on detecting agreement between classifiers belonging to different Machine Learning paradigms. The cases when the majority of the classifiers agreed on predicting a label while the gold standard was different were inspected manually by a specialist in teaching Arabic.
In our Dataset cleaning experiment we used the following classifiers: SVM (with the rbf kernel), Random Forest, KNeighbors, Softmax and XgBoost using linguistic features discussed in Section~\ref{secFeatures}, trained them via cross-validation and compared their majority vote to the gold standard. 

We modified the error classification tags introduced by \citet{di2014multiple} as follows:
\begin{description}[noitemsep]
\item [Wrong] if the classifiers have wrongly labelled the data, and the gold standard is correct.
\item [Modify] if the classifiers are correct and we need to modify the gold standard.
\item [Ambiguous] if we consider both either label is possible based on different perspectives.
\item [False] is an added label which represent the disagreement between the gold standard and the classifiers, when neither is correct.
\end{description} 

For each sentence, five different predictions are assigned. Compared to the gold standard CEFR- label, the classifiers agreed in predicting 10204 instances. Then what we need to consider is when all classifiers agree on the predicted label and it contradicts with the gold standard’s one. In that matter, the classifiers agreed on 1943 sentence classification. We manually investigated random sentences and assigned the error classification tags. We found that the main classification confusion was in Level B instances. The analysis results as in Table 4 show the distribution of categories where each error type occurred. In the end, 380 instances had to be assigned to lower level (usually from B to A).

\subsection{Dataset Two: Simplification examples}

A set of simple/complex parallel sentences has been compiled from the internationally acclaimed Arabic novel \textbf{“Saaq al-Bambuu”} \citep{al2013saud} which has an authorized simplified version for students of Arabic as a second language \citep{al2016saud}. We assume that a successful classifier should be able to detect sentences in the original text that require simplification. Dataset Two consists of 2980 parallel sentences Table ~\ref{Data-Set2}. 

\begin{table}[ht]
\centering
\begin{tabular}{lrl}
\hline \textbf{Level}  & \textbf{Sentence} & \textbf{Token} \\ \hline
Simple A+B  & 2980 & 34447 \\
Complex C  & 2980 & 46521\\
\hline
Total  & 5690 & 80968\\
\hline
\end{tabular}
\caption{\label{Data-Set2} Number of Sentences and Tokens available per each CEFR Level in Dataset two  }
\end{table}

\section{Features and extraction methods}
\label{secFeatures}
We work with following groups of features in Table~\ref{Feature-Set}: Part of speech tagging features (POS-features); Syntactic structure features (Syntactic-features); CEFR-level lexical features; Sentence embeddings. 
\begin{table*}[]
\centering
\begin{tabular}{rlrl}
\hline
\multicolumn{4}{c}{\textbf{POS\_Features}}                                                                            \\
\hline
\textbf{1}  & TTR   of word forms             & \textbf{12}                   & Numeric\_Adj\_Tokens                   \\
\textbf{2}  & Morphemes\_word                 & \textbf{13}                   & Comparative\_Adj\_Tokens               \\
\textbf{3}  & TTR   of Lemma                  & \textbf{14}                   & Conjunction   \_Tokens                 \\
\textbf{4}  & Nouns\_Tokens                   & \textbf{15}                   & Conjunction   \_Subordination \_Tokens \\
\textbf{5}  & Verbs\_Tokens                   & \textbf{16}                   & Proper   noun\_Tokens                  \\
\textbf{6}  & Adj\_Tokens                     & \textbf{17}                   & Pronoun\_Tokens                        \\
\textbf{7}  & Verb\_pseudo\_Tokens            & \textbf{18}                   & Punc\_Tokens                           \\
\textbf{8}  & Passive verbs\_Tokens           & \textbf{19}                   & Simple\_Connector\_Tokens              \\
\textbf{9}  & Perfective   verbs \_Tokens     & \textbf{20}                   & Complex\_   Connector \_Tokens         \\
\textbf{10} & Imperfective verbs   \_Tokens   & \textbf{21}                   & All\_Sent\_ Connector   \_Tokens       \\
\textbf{11} & 3rdperson\_verb\_Verbs & \multicolumn{2}{l}{}                                                   \\
\hline
\multicolumn{4}{c}{\textbf{Syntactic Features}}                                                                        \\
\hline
\textbf{22} & Incidence   of subjects         & \textbf{25}                   & Incidence   of coordination            \\
\textbf{23} & Incidence of objects            & \textbf{26}                   & Average phrases/sentence               \\
\textbf{24} & Incidence   of modifier/root    & \textbf{27}                   & Average   phrases depth                \\
\hline
\multicolumn{4}{c}{\textbf{CEFR Word Features}}                                                                        \\
\hline
\textbf{28} & Incidence of Level A1           & \textbf{32}                   & Incidence of Level C1                  \\
\textbf{29} & Incidence of Level A2           & \textbf{33}                   & Incidence of Level C2                  \\
\textbf{30} & Incidence   of Level B1         & \textbf{34}                   & Word entropy with respect to CEFR\\
\textbf{31} & Incidence   of Level B2         & \multicolumn{1}{l}{\textbf{}} &                                        \\
\hline
\textbf{35} & \multicolumn{3}{c}{\textbf{Sentence Embeddings Features}}   \\                  \hline                      
\end{tabular}
\caption{\label{Feature-Set} The Feature set. (all measures are for the rate of tokens on the sentence levels) }
\end{table*}
\subsection{Linguistic features}
While the sentence-level classification task is novel, we borrowed some features from previous studies of text-level readability \citep{forsyth2014automatic,saddiki2015text,nassiri2018arabic,nassiri2018modern}. We decided to exclude the sentence length from the feature set, as this creates an artificial skew in understanding what is difficult: more difficult writing styles are often associated with longer sentences, but it is not the sentence length which makes them difficult. Specifically, many long Arabic sentences contain shorter ones, which are connected by conjunctions such as ‘\RL{و,} /wa /= and’.  According to the experience of language teachers such sentences do not present problems for the learners.

\subsubsection{The POS-features} 
[Table~\ref{Feature-Set} features (1-21)], these features represent the distribution of different word categories in the sentence, and the morpho-syntactic features of these words. According to \citet{knowles2004notion}, Arabic lemmatization, unlike that of English, is an essential process for analysing Arabic text, because it is a methodology for dictionary construction. Therefore, we used the Lemma/Type ratio instead of Word/Type ratio. Adding features represents the different verb types (Verb pseudo, Passive verbs, Perfective verbs, Imperfective verbs and 3rdperson). As conjunction is one of the important features in representing sentence complexity in Arabic \cite{forsyth2014automatic}, we  used the annotated discourse connectors introduced by \citet{alsaif2012human} by splitting this list into 23 simple connectors and 56 complex connectors referring to non-discourse connectors and discourse connectors respectively. For POS-features extraction we used MADAMIRA a robust Arabic morphological analyser and part of speech tagger \cite{pasha2014madamira}.  
\subsubsection {Syntactic features} 
Features (22-27) from Table~\ref{Feature-Set} provide some information about the sentences structures and number of phrases as well as phases types. These features are derived from a dependency grammar analysis. Because dependency grammar is based on word-word relations, it assumes that the structure of a sentence consists of lexical items that are attached to each other by binary asymmetrical relations, which is known as dependency relations. These relations will be more representative for this task. We used CamelParser \cite{shahrour2016camelparser} a system for Arabic syntactic dependency analysis together with contextually disambiguated morphological features which rely on the MADAMIRA morphological analysis for more robust results. 
\subsubsection {CEFR-level lexical features}  
Features (28-34) from Table~\ref{Feature-Set} are used to assign each word in the sentence with an appropriate CEFR level. For this, we created a new Arabic word list consisting of 8834 unique lemmas labelled with CEFR levels. This list was a combination of three frequency lists, 1) Buckwalter and Parkinson 5000 frequency word list based on 30-million-word corpus of academic/non-academic and written/spoken texts \cite{buckwalter2014frequency} KELLY’s list which is produced from the Kelly project \cite{kilgarriff2014corpus}, which directly mapped a frequency word list to the CEFR levels using numerous corpora and languages, 3) lists presented at the beginning of each chapter in ‘Al-Kitaab’ \cite{brustad2015kitaab}.  Merging the lists and aligning them with the Madamira lemmatiser led to our new wide-coverage Arabic frequency list, which can be used to predict difficulty as Entropy of the probability distribution of each label in a sentence. The current list shows some consistency with the English profile list in terms of the percentage of words allocated to each CEFR level.

\subsection{Sentence embeddings}
In addition to the 34 traditional features we can represent sentences as embedding vectors using different neural models as following:
\paragraph {\emph{fastText}} 
A straightforward way to create sentence representations is to take a weighted average of word embeddings (WE) of each word, for example, using fastText vectors. This embedding was trained on Common Crawl and Wikipedia using fastText \footnote{https://fasttext.cc/docs/en/crawl-vectors.html} tool. Using the Arabic ar.300.bin file in which each word in WE is represented by the 1D vector mapped of 300 attributes \cite{grave2018learning}. We had to normalize the sentence vectors to have the same length with respect to dimensions. For this, we calculated tf-idf weights of each word in the corpus to use them as weights:

\paragraph{$s=w_1 w_2….w_n$}
\begin{equation}
Embed.[s]=\frac{1}{n} \sum_{i}tfidf[w_i]* Embed.[w_i ]
\end{equation}

\paragraph {\emph{Universal sentence encoder}}  \cite{yang2019multilingual}
This model requires modeling the meaning of word sequences rather than just individual words. Also it was generated mainly to be used on the sentence level which after sentence tokenization, it encodes sentence to a 512-dimensional vector. We used here the large version\footnote{https://tfhub.dev/google/universal-sentence-encoder-multilingual/1}.

\paragraph {\emph{Multilingual BERT}} \cite{devlin2018bert}
Pretrained transformers models proved their ability to learn successful representations of language inspired by the transformer model presented in \cite{vaswani2017attention} — who introduced using attention instead to incorporate context information into sequence representation. BERT. Here, we used the last layer produced by BERT transformers while padding the sentences to the maximum length of 128 tokens.  

\paragraph {\emph{XLM-R}} \cite{conneau2019unsupervised}
This is another multilingual BERT-like model, which is different from mBERT by being trained on Common Crawl (instead of Wikipedias) with slightly different parameters. We used the same setup for classification as in the case of mBERT, while also testing a different setup of combining its output with linguistic features and using it as a joined vector of features for traditional ML classification.

\paragraph {\emph{Arabic BERT}} We trained two available BERT-like pre-trained Arabic transformer models available at Hugging face transformers (AraBERT\footnote{https://huggingface.co/aubmindlab/bert-base-arabert} and Arabic-BERT\footnote{https://huggingface.co/asafaya/}).

Both models contain both Modern Standard Arabic (MSA) and Dialectal Arabic (DA). The pre-training data used for the AraBERT model consist of 70 million sentences \citep{antoun2020arabert}. Arabic-BERT trained on both filtered Arabic Common Crawl and a recent dump of Arabic Wikipedia  contain approximately 8.2 Billion words\citep{safaya2020kuisail}.

\section{Experiments}
CEFR language proficiency levels can be presented as labels or as a continuous scale. The former is solved as a classification task with macro-averaged F-1 as the main measure for accuracy. The latter is solved as a regression task \cite{vajjala2014automatic}. At first we decided to work with the three main levels (A,B,and C) because it was quite difficult to determine the boundary between the inner sub levels as in the boundary between B1 and B2.Yet, the other binary classification is either Simple (A+B) or Complex (C). Here there is a problem for evaluation, since the gold standard labels are represented as integers 1, 2, 3 (for the A, B and C levels respectively), which leads to a large number of ties.  Out of the standard correlation measures, Kendall’s tau-b is designed to handle ties, so in addition to Pearson's $\rho$ this is our measure for regression \cite{maurice1990rank}.
\begin{table}[ht]
\centering
\begin{tabular}{lccc}
\textbf{Classification models} & \textbf{P} & \textbf{R} & \textbf{F-1} \\\hline
\textbf{Features} &&&\\\hline
KNeighbors & 0.51 & 0.55 & 0.52 \\
Naive bayes & 0.68 & 0.65 & 0.65 \\
Decision Tree & 0.75 & 0.77 & 0.74 \\
Random Forest & 0.59 & 0.75 & 0.66 \\
XgBoost & 0.74 & 0.77 & 0.74 \\
Softmax & 0.74 & 0.77 & 0.74 \\
SVM, Linear & 0.75 & 0.77 & 0.74 \\
\textbf{SVM, rbf kernel} & \textbf{0.75} & \textbf{0.77} & \textbf{0.75}\\\hline
\textbf{Neural}\\\hline
FastText &0.57&0.59&0.58\\
UCS & 0.52 &0.53  &0.52 \\
mBERT &0.53  & 0.54 &0.53 \\
ArabicBERT &\textbf{0.78}  & \textbf{0.80} & \textbf{0.80}\\
AraBERT & 0.73 &0.73  & 0.73\\
XLM-R & 0.56 & 0.70 & 0.61\\
\end{tabular}

\caption{\label{Classification} 3-way classification using weighted macro-averaged precision, recall and F-1, Dataset One Using all features versus neural models.}
\end{table}

\subsection{Readability as a Classification Problem }
Table ~\ref{Classification} presents the results of classification using updated version of dataset one after application of the error analysis. Applying different ML approaches with 10-fold cross-validation on the 3-way multi-class classification. 
The classification results as presented in Table ~\ref{Classification} divided into two categories: 1) [Linguistics] adding XLM-R vectors to the original set of linguistics features and train with 1058 features [1024 XLM-R dimensions + 34 linguistics features]; 2) [Neural] represents the sentence only by sentence embeddings with neural models.

On the one hand, using linguistic features along with sentence embedding vectors, SVM with rbf kernel classifier provides the best F-1 with 0.75 on the updated corpus version. The SVM classifier is slightly better than both Xgboost and Softmax in precision and they have roughly the same recall value.
On the other hand, comparing sentence embeddings of different kinds such as: XLM-R, mBERT, FasText and UCS along with AraBERT and Arabic-BERT.his indicated that Arabic-BERT is a clear winner with F-1 0.80. Since the architecture for building for all BERT-like models are very similar, we suspect that the more Arabic varied corpus (Common Crawl and Wikipedia for Arabic-BERT vs Common Crawl XML-R vs Wikipedia for BERT, AraBert and UCS) used to train the Arabic-BERT model is responsible for its better performance

The confusion matrix in Table ~\ref{confusion_matrix} shows a clear separation between the lower and higher level of proficiency. The majority of errors are between neighbouring levels and the number of errors decreases when we move away from the true class. The most problematic level was B which has a tendency to be classified as CEFR Level A.

\begin{table}[ht]
\centering
\begin{tabular}{clll}
\hline
\textbf{Predicted} & \textbf{A} & \textbf{B} & \textbf{C} \\\hline
\textbf{A} & 7485 & 1021 & 156 \\
\textbf{B} & 4506 & 1112 & 0 \\
\textbf{C} & 0 & 0 & 8627\\\hline
\end{tabular}
\caption{\label{confusion_matrix} Confusion Matrix of SVM (rbf) on 3-way classification with XLM-R.}
\end{table}

\subsection{Readability as a Regression Problem}
Regression allows us to make ranked predictions along the discrete CEFR levels thus assessing which text is more difficult than the other. The training just as in the previous experiment with applying different ML approaches with 10-fold cross-validation. The results for regression can be rated using mean absolute error (MAE) from the gold standard and the correlation coefficients, Pearson, Spearman and Kendall’s tau. The results are listed in Table ~\ref{Regression}. As with classification, error analysis leads to improved results across all methods. The best MAE rate of 0.34 shows that sentence difficulty prediction is quite close to the gold labels. As mentioned before, our model has a very large number of ties for the gold labels (which can only take three values), so the preferred evaluation measure for regression is Kendall’s tau-b. The best models are RF and SVR on the XLM-R features. 
\begin{table}[ht]
\centering
\setlength{\tabcolsep}{3pt}
\begin{tabular}{lccc}
\hline
\textbf{Model} & \textbf{Pearson} & \textbf{Spearman} & \textbf{Kendall} \\\hline
Decision Tree 2T & 0.82 & 0.62 & 0.44 \\\hline
Decision Tree 5T & \textbf{0.83} & 0.64 & 0.47 \\\hline
Random Forest & \textbf{0.82} & 0.70 & \textbf{0.54} \\\hline
Xgboost & 0.78 & 0.56 & 0.37 \\\hline
Linear & 0.74 & 0.67 & 0.49 \\\hline
MLP & 0.81 & 0.68 & 0.49 \\\hline
SVR,rbf kernel & 0.78 & 0.69 & 0.52 \\\hline
\textbf{SVR, Linear} & 0.8 & \textbf{0.71} & \textbf{0.54}\\\hline
\end{tabular}
\caption{\label{Regression}Regression using all features and XLM-R for sentences}
\end{table}

\subsection{Feature Selection}
Interpreting feature importance and effectiveness is a way for a better understanding of the classification ML model’s logic. This process provides ranking the features by assigning a score for each feature represents its contribution in the target label prediction. These scores provide insights into data representation and model performance. Working with these features ranking can improve the model efficiency and effectiveness by focusing only on the important variables and ignore the irrelevant or noisy features. For this purpose, we applied the Recursive Feature Elimination (RFE) a wrapper method for feature selection approach on the basis of SVM classifier. RFE works with recursively removing some features and testing the remain features to select the best feature set affecting the classifier decisions.
The results of using RFE approach testing the SVM classifier as represented in Table~\ref{Feature-Selection} showing the best ten features contributes to the prediction model. Sentence embedding using XLM-R appeared at the top of the list conveying that it is the most useful feature for sentence difficulty scoring. Followed by the CEFR word frequency features with four features in different positions (Label A1, Label B2, Label C2, and Entropy). The third most effective features are that of the syntactic-set representing more in-depth into the sentence’s syntactic knowledge. 
\begin{table}[ht]
\centering
\begin{tabular}{ll}
\hline
35 & \textbf{Sentence embedding} \\\hline
26 & \textbf{Average phrases/sentence} \\\hline
31 &\textbf {Incidence of Level B2} \\\hline
27 & Average phrases depth \\\hline
28 & \textbf{Incidence of Level A1} \\\hline
24 & Incidence of modifier/root \\\hline
23 & Incidence of objects \\\hline
22 & Incidence of subjects \\\hline
32 & Incidence of Level C1 \\\hline
34 & Words CEFR levels entropy\\\hline
\end{tabular}
\caption{\label{Feature-Selection} List of ten most effective features using REF approach based on SVM classifier }
\end{table}
\subsection{Ablation}
Going further, we performed feature ablation experiments by excluding certain sets of features. We applied SVM rbf classifier on the full dataset while excluding one of the four main group of features blocks POS, Syntactic-features, CEFR-level lexical features, sentence embedding, along with using only the sentence embeddings (XLM-R since this was shown in comparison of embeddings below). This shows that the sentence embeddings significantly contribute to the classification results Table~\ref{Ablation}), in spite of the efforts to create hand-crafted features. Nevertheless, the linguistic features are useful in interpreting the results of purely neural classification. This results prove that the transformer models provide a rich representation for the sentences covering linguistic features.

According to the primary results from the feature selection and the ablation experiments, which proved the use of sentence embedding alone could fulfill the task without extensive use of linguistics features. These results encouraged us to continue experimentation with applying only the sentence embedding feature to reduce the number of features which consequently decreases the data analysis and training time. \begin{table}[ht]
\centering
\begin{tabular}{lccc}
\hline \textbf{Feature set} & \textbf{P} & \textbf{R} & \textbf{F-1} \\\hline
Exclude XLM-R & 0.49 & 0.63 & 0.55 \\\hline
Exclude POS & 0.55 & 0.71 & 0.62 \\\hline
Exclude Syntactic & 0.57 & 0.69 & 0.59 \\\hline
Exclude CEFR & 0.55 & 0.71 & 0.62 \\\hline
\textbf{Only XLM-R} & 0.75 & 0.77 & \textbf{0.75}\\\hline
\end{tabular}
\caption{\label{Ablation} SVM Classification ablation experiment on 3-way classification}
\end{table}

\subsection{Testing on Dataset Two}
For the binary classification, the classifier reached F-1 of 0.94 and 0.98 for Arabic-BERT and SVM XML-R respectively.  However, when testing the binary classifiers trained from DataSet One on Dataset Two the accuracy drops considerably, see Table ~\ref{Testing:1}. 

\begin{table}[t]
\centering
\begin{tabular}{lll}
\hline &\textbf{Arabic-BERT} & \textbf{XLM-R} \\\hline
\textbf{P} & 0.60 & 0.56 \\\hline
\textbf{R} & 0.50 & 0.53 \\\hline
\textbf{F-1} & 0.53 & 0.54\\\hline
\end{tabular}
\caption{\label{Testing:1} Fine-tuned Arabic-BERT versus SVM XLM-R Classifier's performance on Dataset two }
\end{table}

As the confusion matrix in Table~\ref{Testing:Confusion} shows, both classifiers performed better in identifying the complex instances rather than simple ones, so the F1 measure drops. However, the initial results on dataset two shows that XLM-R classifier performed better than Arabic-BERT, we still consider Arabic-BERT classifiers [both 3-way and binary] as best classifier so far. Our interpretation for these confusions is because of the fictional nature of Dataset Two. First, the fiction is well represented in the training data for the A+B levels in Dataset One, while the C level (Snapshot corpus) contains texts of many different types from the internet, so that the classifiers could not handle the mismatch in genres.  The other possible reason is that what is considered as complex sentences which are worth simplification according to developers of  Dataset Two does not really seem to be complex as to be only suitable for the C-level students.  More research is needed to identify the difference between the two datasets.

\begin{table}[t]
\centering
\begin{tabular}{lllll}
\hline &\multicolumn{2}{c}{\textbf{ArabicBert}}&\multicolumn{2}{c}{\textbf{XLM-R}}\\\hline
\textbf{Predicted} & \textbf{A} & \textbf{C} & \textbf{A} & \textbf{C} \\\hline
\textbf{A} &{19} & 2961 & {138} & {2842} \\
\textbf{C} &{46} & 2934 & {223} & {2757}\\\hline
\end{tabular}
\caption{\label{Testing:Confusion} Confusion Matrix with binary classifier Arabic-BERT versus XLM-R on Dataset Two.}
\end{table}

\subsection{Dataset Two Sentence Similarity}
For the purpose of this experiment we needed to include non-simplified sentence to the Dataset two. So that, we duplicated the 2980 complex sentences without simplification aligned with the exact sentence without modification and labeled them with 0 indicating not paraphrased/simplified.  Resulting in a Dataset consist of 2980 sentences with right simplification labeled as 1 and other 2980 non-simplified with label 0, in a total of 5960 sentence.  The two models trained on this similarity task (AraBert and Arabic-Bert) achieve the F-1 measure of 0.98, leading to the ability to detect sentences which need simplification according to the Dataset Two standard.

\section{Related Work on Arabic}
\label{sec:length}
The last two decades have seen enormous efforts (especially for the English language) to develop readability measurement ranging from the traditional readability formulae to ML algorithms. English language researchers have introduced more than 200 readability formulae \citep{dubay2004principles} as well as hundreds of models \citep{schwarm2005reading}. In contrast, less research has addressed Arabic language issues and their challenges for robust readability formulae.\par
Some attempts to formulate statistical formulae for the Arabic language reflected traditional English formulae such as the Flesch–Kincaid Grade. The simplest formulae included the average word length, the average sentence and other surface features. According to \citet{cavalli2018arabic} these simple formulae are Dawood formula (1977), Al-Heeti formula (1984), and the formula presented by \citet{daud2013corpus} based on a corpus. The more sophisticated formulae represent the syllables and more insights the Arabic sentences, such as AARI Base by \citet{al2014aari} and OSMAN by \citet{el2016osman}.\par
Other studies were conducted to measure text readability by targeting either first or second language learners for Arabic language modelled using different ML algorithms. Most of these studies used the previously traditional features along with varying lists of part of speech features (POS) representing the words in each document as in studies by \citep{al2010automatic,forsyth2014automatic,saddiki2015text,nassiri2018arabic}. \citet{forsyth2014automatic} used the word frequency dictionary by \citet{buckwalter2014frequency} to classify the words’ level against this dictionary frequencies. The dictionary was used later by \citet{nassiri2018modern} along with 133 POS features to achieve an accuracy of 100\% with 3-classes. The ‘Al-Kitaab’ textbook has a word list introduced at the beginning of each chapter in the book. These lists were used by \citet{cavalli2014matching} for comparing the words appeared in a text against this list and labelling them by (target, known, unknown). \citet{saddiki2018feature}, highlight adding new syntactic features to their features targeting more in-depth analysis. They used two different datasets for both first and second Arabic language learning. This yielded an accuracy of 94.8\%, 72.4\% for first language learners and second language learners respectively.

\section{Conclusions}
We present the first attempt to build a methodology for Arabic difficulty classification on the \textbf{sentence} level.  We have found that while linguistic features, such as POS tags, syntax or frequency lists are useful for prediction, Deep Learning is the most important contribution to performance, but the traditional features can help in interpreting the black box of Deep Learning alone. For this specific task and for the Arabic language, fine-tuned Arabic-BERT offers better performance than other sentence embedding methods. Also, application of the classifiers trained on one dataset to a very different evaluation corpus shows that the classifiers learn some important properties of what is difficult in Arabic, but the transfer is more successful for the feature-based models than for the BERT-based ones. 

In the end, our best classifier is reasonably reliable in detecting complex sentences; however, it is less successful in separating between the lower learner levels. Still the binary classifier provides the functionality for filtering out really difficult sentences, not suitable for the learners. If we are thinking of Arabic learners especially in higher education, we are expecting learners to graduate with a BA degree in the case of Arabic as a complex language with confidence in reading B2 texts, which implies that the tool for separating A+B vs C level texts is really useful for undergraduate teaching.

Through our tool providing computational assessment of difficulty, we will be able: i) to select the appropriate texts for students; ii) to access ever-larger volumes of information to find educational material of the right difficulty online; iii) to explore curriculum-based assessment to find what is most effective in finding gaps in a curriculum that can be filled according to students’ needs.

Our future work involves building a parallel simple/complex Arabic corpus for sentence simplification. The corpus will be classified on the basis of how difficult the sentences are in a Common Crawl snapshot of Arabic web pages. Using the text difficulty classifier, we can split the corpus into two groups for complex and simple sentences. We also consider the semantic similarity detection on \textbf{“Saaq al-Bambuu”} as a benchmark, which could be used in the corpus compilation.  In this study we only performed some ablation analysis, but because BERT-like models are more useful as the classifiers, we want to investigate their performance via probing for linguistic features following the BERTology framework \cite{rogers20,sharoff21rs}. We also want to explore the link between the difficulty assessment on the document vs sentences levels \cite{dell14assessing}.

\section{Acknowledgments}
This research is a part of PhD project funded by Newton-Mosharafa Fund. All experiments presented in this paper were performed using Advanced Research Computing (ARC) facilities provided by Leeds University.

\bibliography{bibexport}
\bibliographystyle{acl_natbib}

\end{document}